\documentclass[runningheads]{llncs}
\usepackage{graphicx}
\usepackage[T1]{fontenc}
\usepackage{amssymb}
\usepackage{cite}
\usepackage{caption}
\usepackage{footnote} 
\usepackage{bbding}
                       
\usepackage{amsmath}
\usepackage{autobreak}
\usepackage{multirow}

\usepackage{subfigure}
\usepackage{epsfig}
\usepackage{epstopdf}
\usepackage{stfloats}
\usepackage{booktabs}
\usepackage{array}
\usepackage{color}
\usepackage{makecell}
\usepackage{tabularx}
\usepackage{array}
\usepackage{subfig}
\usepackage{booktabs}
\usepackage[table,xcdraw]{xcolor}
\usepackage{makecell}
\usepackage{color}

\usepackage[misc]{ifsym}

\makeatletter
\newcommand*\bigcdot{\mathpalette\bigcdot@{.5}}
\newcommand*\bigcdot@[2]{\mathbin{\vcenter{\hbox{\scalebox{#2}{$\m@th#1\bullet$}}}}}
\makeatother

\begin{document}

\title{ACT-Net: Anchor-context Action Detection in Surgery Videos}

\author{Luoying Hao\inst{1,2,*} \and Yan Hu\inst{2,*,(\textrm{\Letter})} \and Wenjun Lin\inst{2,3} \and Qun Wang\inst{4} \and Heng Li\inst{2} \and Huazhu Fu\inst{5} \and Jinming Duan\inst{1,(\textrm{\Letter})} \and Jiang Liu\inst{2,(\textrm{\Letter})}}

\authorrunning{L.Hao et al.}

\institute{
School of Computer Science, University of Birmingham, UK \\ \email{j.duan@bham.ac.uk} \and%
Research Institute of Trustworthy Autonomous Systems and Dept. of Computer Science and Engineering, Southern University of Science and Technology, China \\ \email{huy3@sustech.edu.cn, liuj@sustech.edu.cn} \\ \and%
Dept. of Mechanical Engineering, National University of Singapore, Singapore \\%
 \and
Third medical center of Chinese PLAGH, China \and%
Institute of High Performance Computing (IHPC), Agency for Science, Technology and Research (A*STAR), Singapore \\ 
*: Luoying Hao and Yan Hu are Co-first authors
 }

\maketitle              
%

\begin{abstract}
Recognition and localization of surgical detailed actions is an essential component of developing a context-aware decision support system. However, most existing detection algorithms fail to provide high-accuracy action classes even having their locations, as they do not consider the surgery procedure's regularity in the whole video. This limitation hinders their application. Moreover, implementing the predictions in  clinical applications seriously needs to convey model confidence to earn entrustment, which is unexplored in surgical action prediction.  In this paper, to accurately detect  fine-grained actions that happen at every moment, we propose an anchor-context action detection network (ACTNet), including an anchor-context detection (ACD) module and a class conditional diffusion (CCD) module, to answer the following questions: 1) where the actions happen; 2) what actions are; 3) how confidence predictions are. Specifically, the proposed ACD module spatially and temporally highlights the regions interacting with  the extracted anchor in surgery video, which outputs action location and its class distribution based on anchor-context interactions. Considering the full distribution of action classes in videos, the CCD module adopts a denoising diffusion-based generative model conditioned on our ACD estimator to further reconstruct accurately the action predictions. Moreover, we utilize the stochastic nature of the diffusion model outputs to access model confidence for each prediction. Our method reports the state-of-the-art performance, with improvements of 4.0\% mAP against baseline on the surgical video dataset.
\keywords{Action detection, Anchor-context, Conditional diffusion, Surgical video}
\end{abstract}

\section{Introduction}
 
Surgery is often an effective therapy that can alleviate disabilities and reduce the risk of death from common conditions \cite{mersh2021clinical}.  While surgical procedures are intended to save lives, errors within the surgery may bring great risks to the patient and even cause sequelae \cite{ nepogodiev2019global}, which emphasizes the development of a computer-assisted system. A context-aware assistant system for surgery can not only  decrease intraoperative adverse events, and enhance the quality of interventional healthcare \cite{vercauteren2019cai4cai}, but also contribute to surgeon training, and assist procedure planning and retrospective analysis\cite{lalys2014surgical}. 

Designing intelligent assistance systems for operating rooms requires an understanding of surgical scenes and procedures \cite{padoy2019machine}. Most current works pay attention to phase and step recognition \cite{twinanda2016endonet,hashimoto2019}, which is to get the major types of events that occurred during the surgery. They merely provided very coarse descriptions of scenes. As the granularity of action increases, the clinical utility  becomes more valuable in providing an accurate depiction of detailed motion \cite{nwoye2022rendezvous,lin2022instrument}. Recent studies focus on fine-grained action recognition by modelling action as a group of the instrument, its role, and its target anatomy and capturing their associations \cite{islam2020learning,seenivasan2022}. Recognizing  targets in different methods is dependent on different surgical scenarios and it also significantly increases the complexity and time consumption for anatomy annotation \cite{zhang2022automatic}. In addition, although most existing methods can provide accurate action positions, the predicted action class is often inaccurate. Moreover, they do not provide any information about the reliability of their output, which is a key requirement for integrating into assistance systems of surgery  \cite{leibig2017leveraging}. Thus, we propose a reliable surgical action detection method in this paper, with high-accuracy action predictions and their confidence.

Mistrust is a major barrier to deep-learning-based predictions applied to clinical implementation \cite{linegang2006human}. Existing works measuring the model uncertainty \cite{lakshminarayanan2017simple,gal2016dropout} often need several-time re-evaluations, and store multiple sets of weights. It is hard for them to apply to surgery assistance applications to get confidence for each prediction directly\cite{lee2023localization}, and they are limited to improving prediction performance. Conditional diffusion-based generative models have received significant attention due to their ability to accurately recover the full distribution of data guided by conditions from the perspective of diffusion probabilistic models  \cite{rombach2022high}. However, they focus on generating high-resolution photo-realistic images. Instead, after observing our surgical video dataset, our conditional diffusion model aims to reconstruct accurately class distribution. We also access the estimation of confidence with the stochastic nature of the diffusion model.

Here, to predict accurately micro-action (fine-grained action) categories happening every moment, we achieve it with two modules. Specifically, a novel anchor-context module for action detection is proposed to highlight the spatio-temporal regions that are interacted with the anchors (we extract instrument features as anchors), which includes surrounding tissues and movement information. Then, with the  constraints of class distributions and the surgical videos, we propose a conditional diffusion model to cover the whole distribution of our data and to accurately reconstruct new predictions based on full learning. Furthermore, our class conditional diffusion model also accesses uncertainty for each prediction, through the stochasticity of outputs.

We summarize our main contributions as follows: 1) We develop an anchor-context action detection network (ACTNet), including an anchor-context detection (ACD) module and a class conditional diffusion (CCD) module, which combines three tasks: i) where actions locate; ii) what actions are; iii) how confident our model is about predictions.  2)  For ACD module, we develop a spatio-temporal anchor interaction block (STAB) to spatially and temporally highlight the context related to the extracted anchor, which provides micro-action location and initial class. 3) By conditioning on the full distribution of action classes in the surgical videos, our proposed class conditional diffusion (CCD) model reconstructs better class prototypes in a stochastic fashion, to provide a more accurate estimations and push the assessment of the model confidence in its predictions. 4) We carry out comparison and ablation study experiments to demonstrate the effectiveness of our proposed algorithm based on cataract surgery.
\begin{figure*}[tb]
\begin{center}
\includegraphics[width=\linewidth]{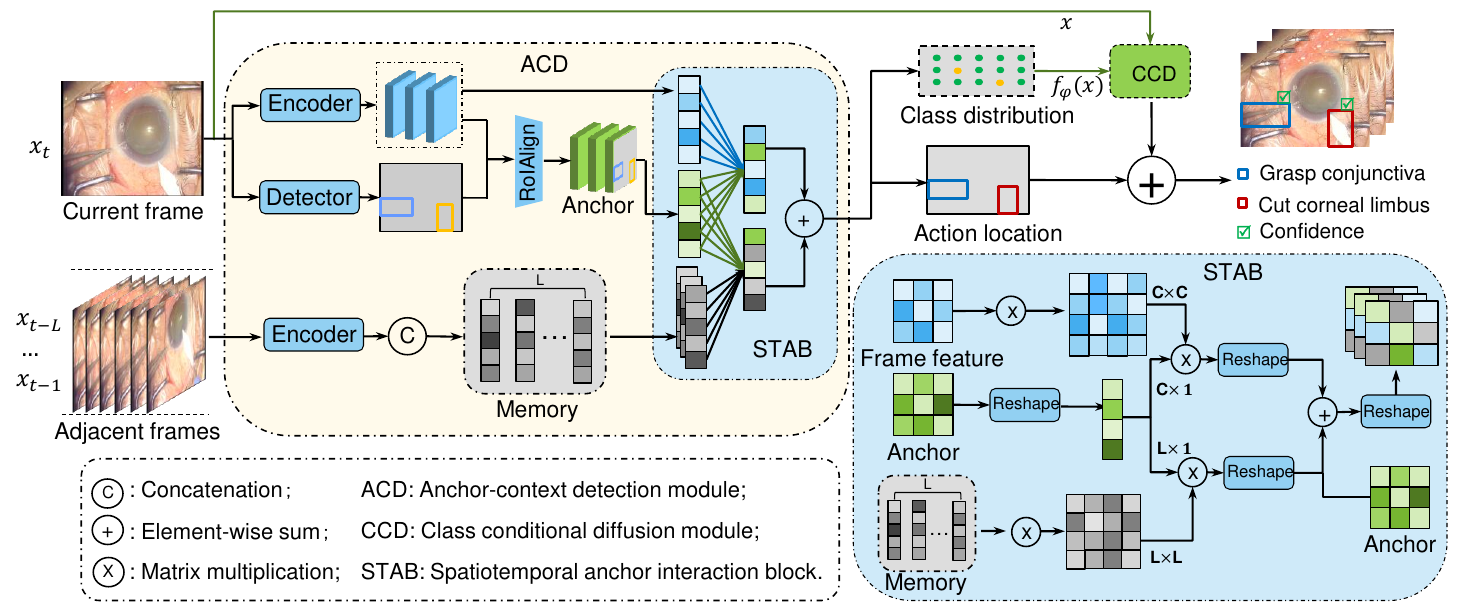}
\end{center}
  \caption{The pipeline of our ACTNet includes ACD and CCD modules.}
\label{fig:pipeline1}
\end{figure*}


\section{Methodology}

The overall framework of our proposed ACTNet for reliable action detection is illustrated in Fig. \ref{fig:pipeline1}. Based on a video frame sequence, the ACD module extracts anchor features and aggregates the spatio-temporal interactions with anchor features by proposed STAB, which generates action locations and initial action class distributions. Then considering the full distribution of action classes in surgical videos, we use the CCD module to refine the action class predictions and access confidence estimations. 

\subsection{ Our ACD module}\label{temporal}

\noindent\textbf{Anchor extraction: } Assuming a video $X$ with $T$ frames, denoted as $X={\{x_t\}}_{t=1}^T$, where $x_t$ is the $t$-th frame of the video. The task of this work is to estimate all potential locations and classes $P={\{box_{n},c_{n}\}}_{n=1}^N$ for action instances contained in video $X$, where $box_{n}$  is the position  of the $n$-th action happened, $c_{n}$ is the action class of  $n$-th action, and $N$ is the number of action instances. For video representation, this work tries to encode the original videos into features based on the backbone ResNet50 \cite{he2016deep} network to get each frame's feature $F=\{f_t\}_{t=1}^T$. 

In surgical videos, the instruments, as action subjects, are  significant to recognize the action. For instrument detection, it is very important but not very complicated. Existing excellent object detection method like Faster R-CNN \cite{ren2015faster} is enough to obtain results with high accuracy. After getting the detected instrument anchors,  RoIAlign is applied to extract the instrument features from frame features. The instrument features are denoted as $I=\{i_t\}_{t=1}^T$. Since multiple instruments exist in surgeries, our action detection needs to solve the problem that related or disparate concurrent actions often lead to wrong predictions. Thus, in this paper, we propose to provide action location and class considering the spatio-temporal anchor interactions in the surgical videos, based on STAB.

\noindent\textbf{Spatio-Temporal Action interaction Block (STAB):} For several actions like pushing, pulling, and cutting, there is no difference just inferred from the local region in one frame. Thus we propose STAB to utilize spatial and temporal interactions with an anchor to improve the prediction accuracy of the action class, which finds actions with strong logical links to provide an accurate class. The structure of STAB is shown in Fig. \ref{fig:pipeline1}. We introduce spatial and temporal interactions respectively in the following.

For spatial interaction: The instrument feature $i_t$ acts as the anchor. In order to improve the anchor features, the module has the ability to select value features that are highly active with the anchor features and merge them. The formulation is defined as: $a_t = \frac{1}{C(f_t)}\sum\limits_{j \in S_j}{h(f_{tj},i_t)g(f_{tj})}$, where $j$ is the index that enumerates all possible positions of $f_t$. A pairwise function $h(\cdot)$ computes the relationship such as affinity between $i_t$ and all $f_{tj}$. In this work, dot-product is employed to compute the similarity. The unary function $g(f_{tj})$ computes a representation of the input signal at the position $j$. The response is normalized by a factor $C(f_t)=\sum\limits_{j \in S_j}{h(f_{tj},i_t)}$. $S_j$ represents the set of all positions $j$. Through the formulation, the output $a_t$ obtains more information from the positions related to the instrument and catches interactions in space for the actions.

\begin{figure*}[tp]
\begin{center}
\includegraphics[width=0.95\linewidth]{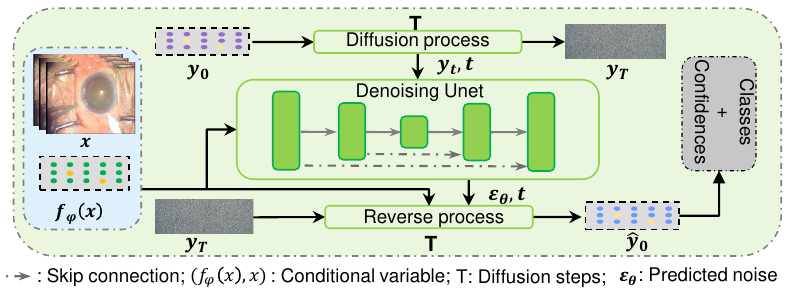}
\end{center}
  \caption{Overview of the class conditional diffusion (CCD) model. }
\label{fig:pipeline2}
\end{figure*}

For temporal interaction: We build memory features consisting of features in consecutive frames: $m_t=[f_{t-L},...,f_{t-1}]$. To effectively model temporal interactions of the anchor, the network offers a powerful tool for capturing the complex and dynamic dependencies that exist between elements in sequential data and anchors. Same with the spatial interaction, we take $i_t$ as an anchor and calculate the interactions between the memory features and the anchor. The formulation is defined as: $b_t = \frac{1}{C(m_t)}\sum\limits_{j \in T_j}{h(m_{tj},i_t)g(m_{tj})}$,where $T_j$ refers to the set of all possible positions along the time series in the range of $L$. In this way, temporal interactions with anchors are obtained. Then a global average pooling is carried out on the spatial and temporal outputs. Action localizations and initial action class distributions are produced based on the fully-connected classifier layer.

\subsection{CCD module for reliable action detection}
Since the surgical procedures follow regularity, we propose a CCD module to reconstruct the action class predictions considering the full distribution of action classes in videos. The diffusion conditioned on the action classes and surgical videos is adopted in our paper. Let $y_0 \in \mathbb{R}^n$ be a sample from our data distribution. As shown in Fig. \ref{fig:pipeline2}, a diffusion model  specified in continuous time  is a generative model with latent $y_t$, obeying a forward process $q_t(y_t|y_{t-1})$ starting at data $y_0$ \cite{ho2020denoising}. $y_0$ indicates a one-hot encoded label vector. 
We treat each one-hot label as a class prototype, i.e., we assume a continuous data and state space,
which enables us to keep the Gaussian diffusion model framework  \cite{ho2020denoising, han2022card}. The forward process and reverse process of unconditional diffusion are provided in the supplementary material.

Here, for the diffusion model optimization can be better guided by meaningful information, we integrate the ACD and our surgical video data as priors or constraints in the diffusion training process. We design a conditional diffusion model $\hat{p}_\theta(y_{t-1}|y_t, x)$ that is conditioned on an additional latent variable $x$. Specifically, the model $\hat{p}_\theta(y_{t-1}|y_t, x)$ is built to approximate the corresponding tractable ground-truth denoising transition step $\hat{p}_t(y_{t-1}|y_t, y_0,x)$. We specify the reverse process with conditional distributions as \cite{pandey2022diffusevae}:
\begin{equation}
\begin{split}
\nonumber
&\hat{p}_t\left(y_{t-1}|y_t,y_0,x\right) = \hat{p}_t\left(y_{t-1}|y_t,y_0,f_\varphi(x)\right) = N\left(y_{t-1};\hat{\mu}\left(y_t,y_0,f_\varphi(x)\right),\hat{\beta}_tI\right)
\label{eq7}
\end{split}
\end{equation}
where $\hat{\mu}\left(y_t,y_0,f_\varphi(x)\right)$ and $\hat{\beta}_t$ are described in supplementary material. $f_\varphi(x)$ is the prior knowledge of the relation between $x$ and $y_0$, i.e., the ACD module pre-trained with our surgical video dataset. The $x$ indicates the input surgical video frames. Since ground-truth step $\hat{p}_t(y_{t-1}|y_t, y_0,x)$ cannot get directly,  the model $\hat{p}_\theta(y_{t-1}|y_t, x)$ are trained by following loss function for estimating $\epsilon_\theta$ to approximate the ground truth:
\begin{equation}
\begin{split}
\nonumber
&\hat{L}(\theta) = {E}_{t,y_0,\epsilon,x}\left[\Arrowvert\epsilon - \epsilon_\theta(x,\sqrt{\Bar{\alpha}_t}y_0+\sqrt{1-\Bar{\alpha}_t}\epsilon+(1-\sqrt{\Bar{\alpha}_t})f_\varphi(x),f_\varphi(x),t)\Arrowvert^2\right]
\label{eq8}
\end{split}
\end{equation}
where $\alpha_t := 1-\beta_t$, $\Bar{\alpha}_t := \prod^t_{s=1}(1-\beta_s)$, $\epsilon \sim N(0,1)$ and $\epsilon_\theta(\cdot)$ estimates $\epsilon$ using a time-conditional network parameterized by $\theta$.  $\beta_t$ is a constant hyperparameter.

To produce model confidence for each action instance, we mainly calculate the prediction interval width (IW). Specifically, we first sample $N$ class prototype reconstruction with the trained diffusion model. Then calculate the IW between the $2.5^{th}$ and $97.5^{th}$ percentiles of the $N$ reconstructed values for all test classes. Compared with traditional classifiers to get deterministic outputs, the denoising diffusion model is a preferable modelling choice due to its ability to produce stochastic outputs, which enables confidence generation.

\section{Experimental Results}
\noindent\textbf{Cataract surgical video Dataset:} To perform reliable action detection, we build a cataract surgical video dataset. Cataract surgery is a procedure to remove the lens of the eyes and, in most cases, replace it with an artificial lens. The dataset consists of 20 videos with a frame rate of 1 fps (a total of 17511 frames and 28426 action instances). Under the direction of ophthalmologists, each video is labelled frame by frame with the categories and locations of the actions. 49 types of action bounding boxes as well as class labels are included in our dataset. The surgical video dataset is randomly split into a training set with 15 videos (13583 frames) and a testing set with 5 videos (3928 frames). 

\noindent\textbf{Implementation Details:}
The proposed architecture is implemented using the publicly available Pytorch Library. A model with ResNet50 backbone from Faster R-CNN-benchmark \cite{ren2016faster} is adopted for our instrument anchor detection. In STAB, we use ten adjacent frames. During inference, detected anchor boxes with a confidence score larger than 0.8 are used.  More implementation details are listed in the supplementary material. The performances are evaluated with official metric frame level mean average precision (mAP) at  IoU = 0.1, 0.3, and 0.5, respectively, obtaining figures in the following  named $mAP_{10}$, $mAP_{30}$ and $mAP_{50}$ with their mean $mAP_{mean}$.
\begin{table}[tp]
\caption{ Methods comparison and ablation study on cataract video dataset.}
\centering
\setlength{\tabcolsep}{1.0mm}{
\begin{tabular}{@{}
>{\columncolor[HTML]{FFFFFF}}l 
>{\columncolor[HTML]{FFFFFF}}c
>{\columncolor[HTML]{FFFFFF}}c
>{\columncolor[HTML]{FFFFFF}}c
>{\columncolor[HTML]{FFFFFF}}c
>{\columncolor[HTML]{FFFFFF}}c 
>{\columncolor[HTML]{FFFFFF}}c 
>{\columncolor[HTML]{FFFFFF}}c 
>{\columncolor[HTML]{FFFFFF}}c 
>{\columncolor[HTML]{FFFFFF}}c 
>{\columncolor[HTML]{FFFFFF}}c @{}}
\toprule
Methods   & $mAP_{10}$     & $mAP_{30}$     & $mAP_{50}$     & $mAP_{mean}$    \\ \midrule
Faster R-CNN \cite{ren2016faster}             & 0.388	& 0.384	& 0.371	& 0.381         \\
 SSD \cite{liu2016ssd}          & 0.360	& 0.358	& 0.350	& 0.356         \\
RetinaNet \cite{lin2017focal}              & 0.358	& 0.356	& 0.347	& 0.354        \\
Mask R-CNN  \cite{he2017mask}             & 0.375	& 0.373	& 0.363	& 0.370        \\
SSDlite  \cite{sandler2018mobilenetv2}            &  0.305	& 0.304	& 0.298	& 0.302      \\
Dynamic R-CNN \cite{zhang2020dynamic}              & 0.315 	& 0.310	& 0.296 	&  0.307     \\
OA-MIL \cite{liu2022robust}             & 0.395  	& 0.394	& 0.378	&  0.389     \\              
\midrule
 backbone             & 0.373		& 0.365		& 0.360		& 0.366         \\
+temporal            & 0.385		& 0.378	& 0.372	& 0.378        \\
+spatial            & 0.394	& 0.385	& 0.377	& 0.385        \\
+STAB              & 0.400	& 0.393	& 0.385	& 0.393        \\
+CCD (ACTNet)             & \textbf{0.415}	& \textbf{0.406}	& \textbf{0.397}	& \textbf{0.406}        \\
\bottomrule
\end{tabular}}
\label{table1}
\end{table}
\begin{figure*}[th]
\begin{center}
\includegraphics[width =\linewidth]{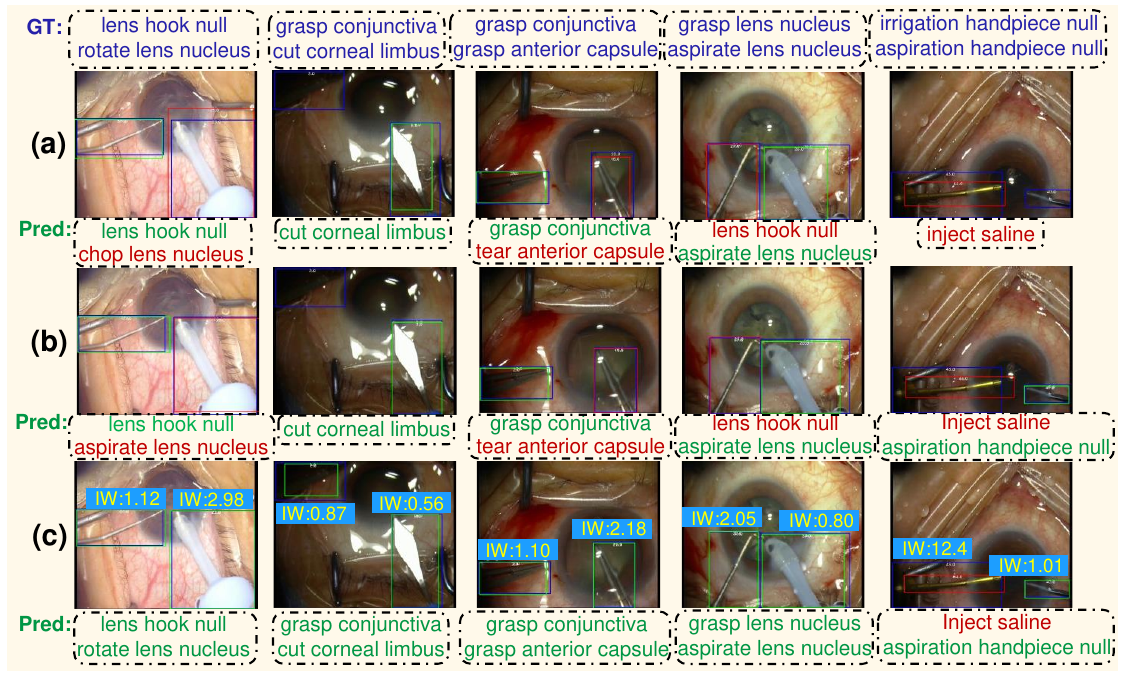}
\end{center}
\caption{Visualization on cataract dataset. We choose different actions to show the results of (a) Faster R-CNN, (b) OA\_MIL, and (c) our ACTNet. For each example, we show the ground-truth (\textcolor{blue}{Blue}), right predictions (\textcolor{green}{Green}) and wrong predictions (\textcolor{red}{Red}). The actions are labelled from left to right. IW values (multiplied by 100) mean prediction interval width to show the level of confidence.}
\label{fig:pipeline3}
\end{figure*}

\noindent\textbf{Method Comparison:}
In order to demonstrate the superiority of the proposed method for surgical action detection, we carry out a comprehensive comparison between the proposed method and the following state-of-the-art methods: 1) single-stage algorithms, including the Single Shot Detector (SSD) \cite{liu2016ssd}, SSDLite \cite{sandler2018mobilenetv2} and RetinaNet \cite{lin2017focal}. 2) two-stage algorithms, including Faster R-CNN \cite{ren2016faster}, Mask R-CNN \cite{he2017mask}, Dynamic R-CNN\cite{zhang2020dynamic} and OA-MIL\cite{liu2022robust}. The data presented in Table \ref{table1} clearly demonstrate that our method outperforms other approaches, irrespective of the IoU threshold being set to 0.1, 0.3, 0.5, or the average values. Notably, the results obtained after incorporating diffusion even surpass Faster R-CNN by 2.5\% and baseline by 4.0\% in terms of average mAP. This finding provides compelling evidence for the efficacy of our method in integrating spatio-temporal interactive 
information under the guidance of anchors and leveraging diffusion to optimize the category distribution. The quantitative results further corroborate the effectiveness of our approach in Fig. \ref{fig:pipeline3}, which  shows that our model does not only improve the performance of the baseline models but also localizes accurately the regions of interest of the actions. More results are listed in the supplementary material.

\noindent\textbf{Ablation Study:}
To validate the effectiveness of our ACTNet, we have done some ablation studies. We train and test the model with spatial interaction, temporal interaction, spatio-temporal interaction (STAB), and finally together with our CCD model. The  testing results are shown in Fig. \ref{fig:pipeline3} and Table. \ref{table1}.  For our  backbone, it is achieved by concatenating the anchor features through RoIAlign and the corresponding frame features to get the detected action classes.

The results reveal that the spatial and temporal interactions for instruments can provide useful information to detect the actions. What's more, spatial interaction has slightly better performance than temporal interaction. It may be led by the number of spatially related action categories being slightly more than that of temporally related action categories. It is worth noting that spatial interaction and temporal interaction can be enhanced by each other and achieve optimal performance. After being enhanced by the diffusion model conditioned on our obtained class distributions and video frames, we get optimal performance.

\noindent\textbf{Confidence analysis:}
To analyze the model confidence, we take the best prediction for each instance to calculate the instance accuracy. We can observe from Table \ref{table2} across the test set, the mean\_IW  of the class label among correctly classified instances by ACTNet is significantly narrower compared to that of incorrectly classified instances. This observation indicates that the model is more confident in its accurate predictions and is more likely to make errors when its predictions are vague. Furthermore, upon comparing the mean\_IW at the true class level, we find that a more precise class tends to exhibit a larger disparity between the correct and incorrect predictions. Fig. \ref{fig:pipeline3} also shows the confidence estimations for some samples. We can see the correct prediction gets smaller IW values compared with the incorrect one (The rightmost figure in column (c)), which means it has more uncertainty for the incorrect prediction.
\begin{table}[thbp]
\caption{ The mean\_IW (multiplied by 100) results from our CCD module.}
\centering
\setlength{\tabcolsep}{1.0mm}{
\begin{tabular}{@{}
>{\columncolor[HTML]{FFFFFF}}l 
>{\columncolor[HTML]{FFFFFF}}c
>{\columncolor[HTML]{FFFFFF}}c
>{\columncolor[HTML]{FFFFFF}}c
>{\columncolor[HTML]{FFFFFF}}c
>{\columncolor[HTML]{FFFFFF}}c 
>{\columncolor[HTML]{FFFFFF}}c 
>{\columncolor[HTML]{FFFFFF}}c 
>{\columncolor[HTML]{FFFFFF}}c 
>{\columncolor[HTML]{FFFFFF}}c 
>{\columncolor[HTML]{FFFFFF}}c @{}}
\toprule
Class   & Instance   & Accuracy    & Mean\_IW (Correct)     & Mean\_IW (Incorrect)        \\ \midrule
grasp conjunctiva              &  487	& 0.702	& 0.91 (342)	& 9.78 (145)     \\
aspirate lens cortex             & 168	& 0.613	& 1.37 (103)	& 15.82 (65)        \\
chop lens nucleus             & 652	& 0.607	& 0.54 (396)	& 9.73 (256)        \\
polish intraocular lens             & 222	& 0.572	& 0.90 (127)	& 8.48 (95)        \\
aspirate lens nucleus             & 621	& 0.554	& 0.76 (344)	& 10.30 (277)        \\
inject viscoelastic               & 112	& 0.536	& 2.17 (60)	& 9.14 (52)        \\
Remove lens cortex               & 174	& 0.471	& 0.42 (82)	& 5.84 (92)        \\
forceps null            & 280	& 0.464	& 2.67 (130)	& 8.38 (150)         \\
\bottomrule
\end{tabular}}
\label{table2}
\end{table}
\section{Conclusions}

In this paper, we propose a conditional diffusion-based anchor-context spatio-temporal action detection network (ACTNet) to achieve recognition and localization of every occurring action in the surgical scenes. ACTNet improves the accuracy of the predicted action class from two considerations, including spatio-temporal interactions with anchors by the proposed STAB and full distribution of action classes by class conditional diffusion (CCD) module, which also provides uncertainty in surgical scenes. Experiments based on cataract surgery demonstrate the effectiveness of our method. Overall, the proposed ACTNet presents a promising avenue for improving the accuracy and reliability of action detection in surgical scenes.
\section{Acknowledgement}
This work was supported in part by General Program of National Natural Science Foundation of China (82272086 and 82102189), Guangdong Basic and Applied Basic Research Foundation (2021A1515012195), Shenzhen Stable Support Plan Program (20220815111736001 and 20200925174052004), and Agency for Science, Technology and Research (A*STAR) Advanced Manufacturing and Engineering (AME) Programmatic Fund (A20H4b0141) and Central Research Fund (CRF).

\bibliographystyle{splncs04}
\bibliography{paper1160_reference.bib}

\begin{thebibliography}{10}
\providecommand{\url}[1]{\texttt{#1}}
\providecommand{\urlprefix}{URL }
\providecommand{\doi}[1]{https://doi.org/#1}

\bibitem{gal2016dropout}
Gal, Y., Ghahramani, Z.: Dropout as a bayesian approximation: Representing
  model uncertainty in deep learning. In: international conference on machine
  learning. pp. 1050--1059. PMLR (2016)

\bibitem{han2022card}
Han, X., Zheng, H., Zhou, M.: Card: Classification and regression diffusion
  models. arXiv preprint arXiv:2206.07275  (2022)

\bibitem{hashimoto2019}
Hashimoto, D.A., Rosman, G., Witkowski, E.R., Stafford, C., Navarette-Welton,
  A.J., Rattner, D.W., Lillemoe, K.D., Rus, D.L., Meireles, O.R.: Computer
  vision analysis of intraoperative video: Automated recognition of operative
  steps in laparoscopic sleeve gastrectomy. Annals of surgery  (2019)

\bibitem{he2017mask}
He, K., Gkioxari, G., Doll{\'a}r, P., Girshick, R.: Mask r-cnn. In: Proceedings
  of the IEEE international conference on computer vision. pp. 2961--2969
  (2017)

\bibitem{he2016deep}
He, K., Zhang, X., Ren, S., Sun, J.: Deep residual learning for image
  recognition. In: Proceedings of the IEEE conference on computer vision and
  pattern recognition. pp. 770--778 (2016)

\bibitem{ho2020denoising}
Ho, J., Jain, A., Abbeel, P.: Denoising diffusion probabilistic models.
  Advances in Neural Information Processing Systems  \textbf{33},  6840--6851
  (2020)

\bibitem{islam2020learning}
Islam, M., Seenivasan, L., Ming, L.C., Ren, H.: Learning and reasoning with the
  graph structure representation in robotic surgery. In: Medical Image
  Computing and Computer Assisted Intervention--MICCAI 2020: 23rd International
  Conference, Lima, Peru, October 4--8, 2020, Proceedings, Part III 23. pp.
  627--636. Springer (2020)

\bibitem{lakshminarayanan2017simple}
Lakshminarayanan, B., Pritzel, A., Blundell, C.: Simple and scalable predictive
  uncertainty estimation using deep ensembles. Advances in neural information
  processing systems  \textbf{30} (2017)

\bibitem{lalys2014surgical}
Lalys, F., Jannin, P.: Surgical process modelling: a review. International
  journal of computer assisted radiology and surgery  \textbf{9},  495--511
  (2014)

\bibitem{lee2023localization}
Lee, Y., Hwang, J.w., Kim, H.I., Yun, K., Kwon, Y., Bae, Y., Hwang, S.J.:
  Localization uncertainty estimation for anchor-free object detection. In:
  Computer Vision--ECCV 2022 Workshops: Tel Aviv, Israel, October 23--27, 2022,
  Proceedings, Part VIII. pp. 27--42. Springer (2023)

\bibitem{leibig2017leveraging}
Leibig, C., Allken, V., Ayhan, M.S., Berens, P., Wahl, S.: Leveraging
  uncertainty information from deep neural networks for disease detection.
  Scientific reports  \textbf{7}(1),  1--14 (2017)

\bibitem{lin2017focal}
Lin, T.Y., Goyal, P., Girshick, R., He, K., Doll{\'a}r, P.: Focal loss for
  dense object detection. In: Proceedings of the IEEE international conference
  on computer vision. pp. 2980--2988 (2017)

\bibitem{lin2022instrument}
Lin, W., Hu, Y., Hao, L., Zhou, D., Yang, M., Fu, H., Chui, C., Liu, J.:
  Instrument-tissue interaction quintuple detection in surgery videos. In:
  Medical Image Computing and Computer Assisted Intervention--MICCAI 2022: 25th
  International Conference, Singapore, September 18--22, 2022, Proceedings,
  Part VII. pp. 399--409. Springer (2022)

\bibitem{linegang2006human}
Linegang, M.P., Stoner, H.A., Patterson, M.J., Seppelt, B.D., Hoffman, J.D.,
  Crittendon, Z.B., Lee, J.D.: Human-automation collaboration in dynamic
  mission planning: A challenge requiring an ecological approach. In:
  Proceedings of the human factors and ergonomics society annual meeting.
  vol.~50, pp. 2482--2486. SAGE Publications Sage CA: Los Angeles, CA (2006)

\bibitem{liu2022robust}
Liu, C., Wang, K., Lu, H., Cao, Z., Zhang, Z.: Robust object detection with
  inaccurate bounding boxes. In: Computer Vision--ECCV 2022: 17th European
  Conference, Tel Aviv, Israel, October 23--27, 2022, Proceedings, Part X. pp.
  53--69. Springer (2022)

\bibitem{liu2016ssd}
Liu, W., Anguelov, D., Erhan, D., Szegedy, C., Reed, S., Fu, C.Y., Berg, A.C.:
  Ssd: Single shot multibox detector. In: Computer Vision--ECCV 2016: 14th
  European Conference, Amsterdam, The Netherlands, October 11--14, 2016,
  Proceedings, Part I 14. pp. 21--37. Springer (2016)

\bibitem{mersh2021clinical}
Mersh, A.T., Melesse, D.Y., Chekol, W.B.: A clinical perspective study on the
  compliance of surgical safety checklist in all surgical procedures done in
  operation theatres, in a teaching hospital, ethiopia, 2021: A clinical
  perspective study. Annals of Medicine and Surgery  \textbf{69},  102702
  (2021)

\bibitem{nepogodiev2019global}
Nepogodiev, D., Martin, J., Biccard, B., Makupe, A., Bhangu, A., Ademuyiwa, A.,
  Adisa, A.O., Aguilera, M.L., Chakrabortee, S., Fitzgerald, J.E., et~al.:
  Global burden of postoperative death. The Lancet  \textbf{393}(10170), ~401
  (2019)

\bibitem{nwoye2022rendezvous}
Nwoye, C.I., Yu, T., Gonzalez, C., Seeliger, B., Mascagni, P., Mutter, D.,
  Marescaux, J., Padoy, N.: Rendezvous: Attention mechanisms for the
  recognition of surgical action triplets in endoscopic videos. Medical Image
  Analysis  \textbf{78},  102433 (2022)

\bibitem{padoy2019machine}
Padoy, N.: Machine and deep learning for workflow recognition during surgery.
  Minimally Invasive Therapy \& Allied Technologies  \textbf{28}(2),  82--90
  (2019)

\bibitem{pandey2022diffusevae}
Pandey, K., Mukherjee, A., Rai, P., Kumar, A.: Diffusevae: Efficient,
  controllable and high-fidelity generation from low-dimensional latents. arXiv
  preprint arXiv:2201.00308  (2022)

\bibitem{ren2015faster}
Ren, S., He, K., Girshick, R., Sun, J.: Faster r-cnn: Towards real-time object
  detection with region proposal networks. Advances in neural information
  processing systems  \textbf{28} (2015)

\bibitem{ren2016faster}
Ren, S., He, K., Girshick, R., Sun, J.: Faster r-cnn: towards real-time object
  detection with region proposal networks. IEEE transactions on pattern
  analysis and machine intelligence  \textbf{39}(6),  1137--1149 (2016)

\bibitem{rombach2022high}
Rombach, R., Blattmann, A., Lorenz, D., Esser, P., Ommer, B.: High-resolution
  image synthesis with latent diffusion models. 2022 ieee. In: CVF Conference
  on Computer Vision and Pattern Recognition (CVPR). pp. 10674--10685 (2022)

\bibitem{sandler2018mobilenetv2}
Sandler, M., Howard, A., Zhu, M., Zhmoginov, A., Chen, L.C.: Mobilenetv2:
  Inverted residuals and linear bottlenecks. In: Proceedings of the IEEE
  conference on computer vision and pattern recognition. pp. 4510--4520 (2018)

\bibitem{seenivasan2022}
Seenivasan, L., Mitheran, S., Islam, M., Ren, H.: Global-reasoned multi-task
  learning model for surgical scene understanding. IEEE Robotics and Automation
  Letters pp.~1--1 (2022). \doi{10.1109/LRA.2022.3146544}

\bibitem{twinanda2016endonet}
Twinanda, A.P., Shehata, S., Mutter, D., Marescaux, J., De~Mathelin, M., Padoy,
  N.: Endonet: a deep architecture for recognition tasks on laparoscopic
  videos. IEEE transactions on medical imaging  \textbf{36}(1),  86--97 (2016)

\bibitem{vercauteren2019cai4cai}
Vercauteren, T., Unberath, M., Padoy, N., Navab, N.: Cai4cai: the rise of
  contextual artificial intelligence in computer-assisted interventions.
  Proceedings of the IEEE  \textbf{108}(1),  198--214 (2019)

\bibitem{zhang2020dynamic}
Zhang, H., Chang, H., Ma, B., Wang, N., Chen, X.: Dynamic r-cnn: Towards high
  quality object detection via dynamic training. In: Computer Vision--ECCV
  2020: 16th European Conference, Glasgow, UK, August 23--28, 2020,
  Proceedings, Part XV 16. pp. 260--275. Springer (2020)

\bibitem{zhang2022automatic}
Zhang, J., Shi, S., Wang, Y., Wan, C., Zhao, H., Cai, X., Ding, H.: Automatic
  keyframe detection for critical actions from the experience of expert
  surgeons. In: 2022 IEEE/RSJ International Conference on Intelligent Robots
  and Systems (IROS). pp. 8049--8056. IEEE (2022)

\end{thebibliography}
\end{document}